\documentclass[letterpaper, 10 pt, conference]{ieeeconf}  

\IEEEoverridecommandlockouts                              

\usepackage{graphics} 
\usepackage{amsmath} 
\usepackage{amssymb}  
\usepackage{kotex}
\usepackage{booktabs}
\usepackage{multirow}

\usepackage{graphicx}

\title{\LARGE \bf
A Dual-Cycled Cross-View Transformer Network for Unified Road Layout Estimation and 3D Object Detection in the Bird's-Eye-View
}

\author{Curie Kim and Ue-Hwan Kim$^{*}$
\thanks{This work was supported in part by the Institute for Information \& communications Technology Promotion (IITP) grant funded by the Korea government (MSIT) (No.2022-0-00907, Development of AI Bots Collaboration Platform and Self-organizing AI, No.2019-0-01842, Artificial Intelligence Graduate School Program (GIST)) and the National Research Foundation of Korea (NRF) grant funded by the Korea government (MSIT) (No. NRF-2022R1C1C1009989).}
\thanks{The authors are with the AI Graduate School, GIST (Gwang-ju Institute of Science and Technology), Gwang-ju, 61005, Repulic of Korea (e-mail: {\tt\small \{curie3170, uehwan\}@gist.ac.kr).}}%
\thanks{*Corresponding author.}%
}

\begin{document}

\maketitle
\thispagestyle{empty}
\pagestyle{empty}

\begin{abstract}

The bird's-eye-view (BEV) representation allows robust learning of multiple tasks for autonomous driving including road layout estimation and 3D object detection. However, contemporary methods for unified road layout estimation and 3D object detection rarely handle the class imbalance of the training dataset and multi-class learning to reduce the total number of networks required. To overcome these limitations, we propose a unified model for road layout estimation and 3D object detection inspired by the transformer architecture and the CycleGAN learning framework. The proposed model deals with the performance degradation due to the class imbalance of the dataset utilizing the focal loss and the proposed dual cycle loss. Moreover, we set up extensive learning scenarios to study the effect of multi-class learning for road layout estimation in various situations. To verify the effectiveness of the proposed model and the learning scheme, we conduct a thorough ablation study and a comparative study. The experiment results attest the effectiveness of our model; we achieve state-of-the-art performance in both the road layout estimation and 3D object detection tasks.

\end{abstract}

\section{Introduction}
Recently, a number of prominent research outcomes are leading the advancement of autonomous driving: 3D object detection \cite{jiang2021vic}, semantic segmentation \cite{kumaar2021cabinet}, and visual odometry \cite{kim2022simvodis++}. Among them, the task of concurrent semantic road layout estimation and 3D object detection in the bird's-eye-view (BEV) \cite{saha2022translating} is attracting growing interest; the low dimensional representation of the surrounding environments enhances crucial tasks for autonomous driving such as navigation and hazard avoidance in addition to the road layout estimation and 3D object detection tasks.

Nonetheless, contemporary methods for concurrent road layout estimation and 3D object detection in BEV display two major limitations. First, contemporary methods hardly consider the imbalance in the class distribution of the data for road layout estimation and 3D object detection. They simply learn to estimate the road layout and the 3D objects in a scene with naive objective functions. Next, contemporary approaches require multiple networks for each class for road layout estimation, resulting in computational inefficiency. Rather than learning one model for multiple classes, these approaches learn separate models for each class (e.g., road, vehicle, pedestrian and background).

To overcome the limitations of contemporary methods, we design a unified model for road layout estimation and 3D object detection inspired by the transformer architecture \cite{vaswani2017attention} and the CycleGAN learning framework \cite{zhu2017unpaired}: the dual-cycled cross-view transformer (DCT) architecture. First of all, the proposed model learns to estimate the road layout and detect 3D objects in a scene taking the class imbalance of the training dataset into account. Specifically, we propose to learn the model based on the focal loss \cite{lin2017focal} so that it focuses on a sparse set of hard examples while restraining the large number of easy negatives from overpowering the training procedure. Moreover, we propose a dual cycle loss to further compensate for the performance degradation caused by the class imbalance in the dataset.

Furthermore, we thoroughly investigate the effect of multi-class learning on the road layout estimation task; since multi-class learning for road layout estimation could reduce the total number of neural networks, it plays a crucial role in autonomous driving where computational resources are restricted. To accomplish this goal, we construct comprehensive multi-class learning experiment scenarios for road layout estimation including the proposed learning scheme (the dual cycle loss), examine the multi-class learning experiment results meticulously, and reveal essential insights for a future research direction.

In summary, the main contributions of our work are as follows:
\begin{enumerate}
    \item \textbf{DCT Architecture}: We propose the dual-cycled cross-view transformer (DCT) network for unified road layout estimation and 3D object detection for autonomous driving along with the learning scheme to handle the class imbalance.
    \item \textbf{Multi-Class Learning}: We investigate the effect of multi-class learning in the context of road layout estimation for the first time to the best of our knowledge.
    \item \textbf{Ablation Study}: We conduct a thorough ablation study and reveal important intuitions for the effect of each design choice.
    \item \textbf{SoTA Performance}: We achieve state-of-the-art performance on both road layout estimation and 3D object detection in the Argoverse and KITTI 3D Object datasets, respectively.
    \item \textbf{Open Source}: We contribute to the research society by making the source code of the proposed DCT network and the pretrained network parameters public.
\end{enumerate}

\section{Related Works}
In this section, we review previous methods relevant to the semantic road layout estimation and monocular 3D object detection tasks in BEV.

\textbf{Semantic road layout estimation in BEV.}
With the advancement of large-scale datasets for autonomous driving \cite{geiger2012we, caesar2020nuscenes, houston2021one}, the task of semantic road layout estimation has emerged. Early approaches have assumed that scenes are in general planar in the autonomous driving settings and transformed images into BEV using simple homographies \cite{sengupta2012automatic, garnett20193d, kim2019deep}. However, the planar scene assumption results in artifacts for dynamic objects such as moving vehicles and pedestrians. Thus, contemporary approaches directly learn the image-to-BEV transformation \cite{saha2022translating, pan2020cross}. For example, VED utilizes a variational autoencoder (VAE) model for producing the semantic road layout of the given image \cite{lu2019monocular}; VPN performs cross-view semantic segmentation by learning common feature representation across multiple views \cite{pan2020cross}; and lift-splat-shoot (LSS) transforms 2D image features into 3D space and generates BEV grids \cite{philion2020lift}.

\textbf{Monocular 3D object detection.}
Monocular 3D object detection aims to estimate the dimension and the orientation of objects in the real-world coordinate from a single monocular image. One category of approaches formulates monocular 3D object detection as 2D object detection and evaluates the depth of the given monocular scene \cite{lu2019monocular, wang2019monocular}. Moreover, approaches based on psuedo-lidars first predict the depth map of the given 2D image, project the predicted depth map into 3D points and conduct 3D object detection directly on these projected 3D points \cite{wang2019pseudo, ma2020rethinking} using 3D object detection algorithms \cite{lang2019pointpillars, pan20213d}.

On the other hand, formulating 3D object detection as 2D segmentation in the top-view is drawing increasing attention due to the conciseness and diverse applicability of the low-dimensional representation in the context of autonomous driving \cite{li2019gs3d, srivastava2019learning, wang2019monocular}. For instance, Mono3D predicts 3D bounding boxes on the ground plane and projects the predicted 3D bounding boxes into the given monocular image \cite{chen2016monocular}; and OFTNet transforms the given monocular image into the top-view representation and performs 2D semantic segmentation for 3D object detection \cite{roddick2019orthographic}. Recently Monolayout \cite{mani2020monolayout} and CVT \cite{yang2021projecting} have attempted to present a unified model for both the semantic road layout estimation and 3D object detection tasks in BEV from a single monocular image.

\textbf{Transformers for computer vision.}
Currently, the transformer architecture \cite{vaswani2017attention} is prevailing in a lot of computer vision fields \cite{zheng2022automatic, zhang2022patchformer}. Based on the attention mechanism first proposed for machine translation \cite{bahdanau2014neural}, transformers explicitly model pair-wise interactions of elements in a sequence. This ability of transformers could aid modeling the correlation between the features of views in different domains \cite{yun2022improving} or in different time-steps \cite{chen2021transformer}. For semantic road layout estimation and 3D object detection in BEV, CVT has designed a cross-view transformer for establishing the correlation between the front-view and the top-view \cite{yang2021projecting}.
\section{Methodology}
\begin{figure*}[t]
\begin{center}
   \includegraphics[width=0.95\linewidth]{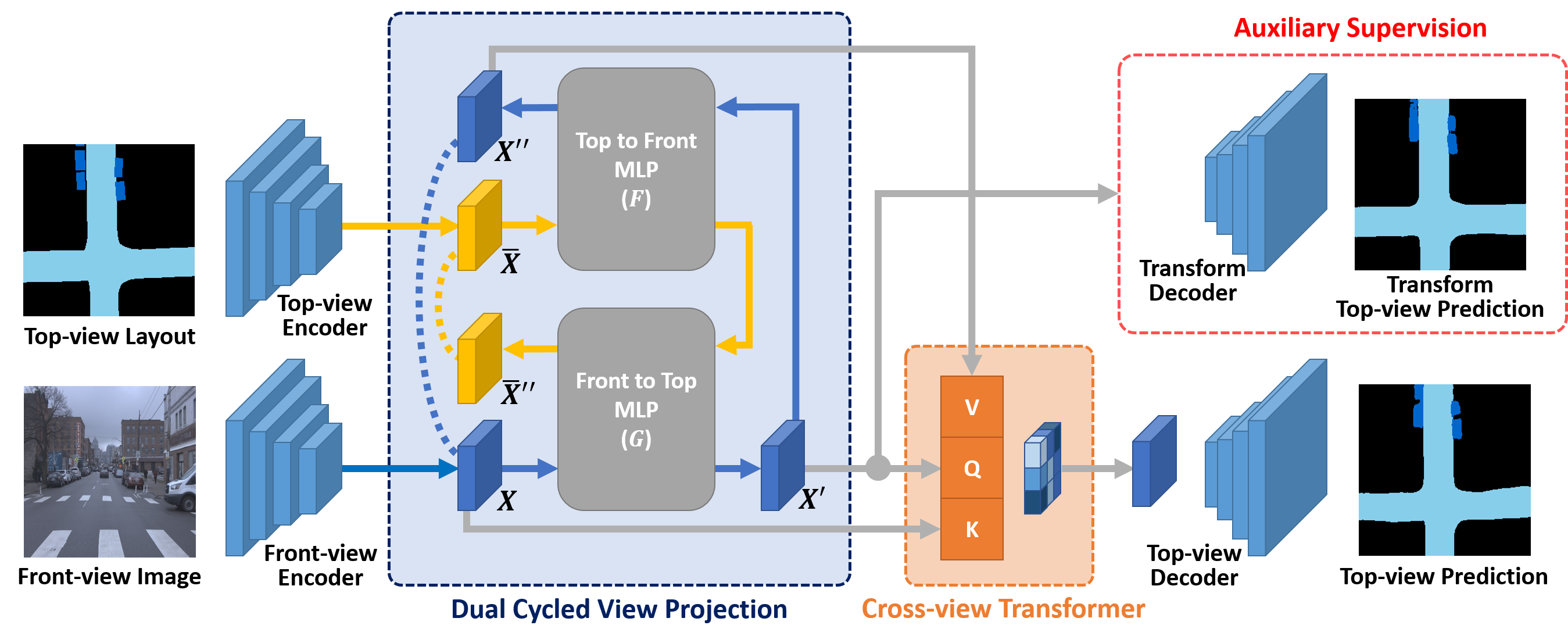}
\end{center}
   \caption{The proposed dual-cycled cross-view transformer (DCT) architecture. The DCT network requires both the top-view layout and the front-view images for training; these two inputs get transformed to another feature representation for the dual cycle loss. When deployed, the DCT network receives just a front-view image to predict the road layout and detect objects.}
\label{fig:overview}
\end{figure*}

We describe the proposed dual-cycled cross-view transformer (DCT) architecture inspirited by the transformer architecture \cite{vaswani2017attention} and the CycleGAN learning framework \cite{zhu2017unpaired} in this section. 

\subsection{Model Overview}
Fig. \ref{fig:overview} depicts the overall architecture of the proposed dual-cycled cross-view transformer (DCT) network. During training, the DCT network receives both the top-view layout and the front-view image as input. The top-view layout and the front-view image get transformed to the front-view representation and the top-view representation, respectively. Then, the transformed front-view embedding and the transformed top-view embedding get cyclically retransformed into the top-view representation and the front-view representation, respectively. The proposed dual-cycled view projection module utilizes these transformed representations for ensuring the complete dual cycle consistency---leading to performance enhancement. Further, the top-view representation extracted from the input front-view image goes through the cross-view transformer \cite{yang2021projecting} and then becomes the top-view prediction, i.e., the semantic road layout from the single front-view image. In addition, we utilize an auxiliary supervision for better gradient flow during training. During testing or inference, the DCT network only receives a front-view image for road layout estimation and 3D object detection with the cross-view transformer.
\subsection{Remedy for Class Imbalance}
When the distribution of class labels in the training set is not even but skewed, the vast number of easy negative samples tend to overpower the training procedure and the model learns not meaningful representation with the naive cross-entropy loss \cite{lin2017focal}. In the road layout estimation task, the layouts ordinarily contain way more background pixels than other classes, i.e., vehicles or roads. Thus, an appropriate management of the class imbalance is essential and it leads to significant performance enhancement.

To attain the goal, we propose to learn road layout estimation and 3D object detection in BEV from a single front-view image utilizing the focal loss \cite{lin2017focal} as follows:

\begin{equation}\label{focal_loss}
    L_{focal} = \sum_{c=1}^{N}(1-p_{c})^\gamma \log(p_{c}),
\end{equation}
\noindent where $p_{c} = log\frac{exp(x_{c})}{\sum_{i=1}^{C}exp(x_{i})} \in [0 , 1]$, $N$ and $\gamma$ represent the estimated probability of the class $c$, the total number of class categories and a hyper-parameter, respectively. In the single class learning setting, $c \in \{\text{foreground}, \text{background}\}$ and we set $\gamma$ as 2.

Particularly, we employ the focal loss in two stages of the proposed learning scheme for the top-view prediction (road layout estimation and 3D object detection): 1) the auxiliary supervision and 2) the top-view prediction using the cross-view transformer.

\subsection{Auxiliary Supervision}
Neural architectures that incorporate multiple components could bring about a large number of parameters---making the optimization process challenging \cite{szegedy2015going}. To cope with a possible gradient vanishing problem and better flow the supervision signal throughout the entire architecture, we employ an auxiliary supervision branch. The auxiliary supervision branch takes $X'$ from the dual cycled view projection module and predicts the road layout from $X'$ using a separate decoder (the transform decoder).
\subsection{Dual Cycle Loss} \label{Dual Cycled View Projection}
Inspired by the CycleGAN learning framework \cite{zhu2017unpaired}, we propose the dual cycled view projection module which evaluates the dual cycle loss. First, the proposed dual cycled view projection module consists of four major components: a front-view feature extractor (i.e., encoder), a top-view feature extractor, a top-to-front multi-layer perceptron (MLP) and a front-to-top MLP. The two feature extractors learn representations for top-view layouts and front-view images. Moreover, the two MLPs conduct domain transfer from top-to-front and front-to-top.

Next, the proposed dual cycle loss boosts the performance of the view transformation by letting the DCT network learn both the forward view transformation (front-view to top-view to front-view) and the backward view transformation (top-view to front-view to top-view); the dual cycle loss comprise of a forward loss and a backward loss.

Furthermore, for mathematical description of the dual cycle loss, let $X$, $\bar{X}$, $F$ and $G$ denote the front image embedding (i.e., the output of the front-view feature extractor), the top-view layout embedding (i.e., the output of the top-view feature extractor), the top-to-front MLP and the front-to-top MLP, respectively. Then, $X^{\prime} = G(X)$, $X^{\prime\prime} = F(X^{\prime}) = F(G(X))$, and $\bar{X}^{\prime\prime}=G(F(\bar{X}))$ lie in the top-view, the front-view and the top-view embedding spaces, respectively. Consequently, the forward loss becomes

\begin{equation}
    L_{fw} = \|F(G(X))-X\|_{1} = \|X''-X\|_{1},
\end{equation}

\noindent and similarly the backward loss is

\begin{equation}
  L_{bw} = \|G(F(\bar{X})) - \bar{X}\|_{1} = \|\bar{X}^{\prime\prime} - \bar{X}\|_{1}.
\end{equation}

\noindent Finally, we define the dual cycle loss as follows:
\begin{align}
\begin{split}
    L_{dual-cycle} &= L_{fw} + L_{bw} \\
                   &= \| X^{\prime\prime} - X\|_{1} + \| \bar{X}^{\prime\prime} - \bar{X}\|_{1}.
\end{split}
\end{align}

\subsection{Multi-Class Learning}

Learning cross-view transformation of each entity (vehicles and roads) using separate neural networks results in a substantial computational inefficiency as practiced in conventional methods \cite{mani2020monolayout, yang2021projecting}. Thus, multi-class learning would hugely improve the computational efficiency for autonomous driving systems where the computational capacity is in general limited. To investigate the effect of multi-class learning in the unified road layout estimation and 3D object detection learning environments, we use $c \in \{\text{vehicle}, \text{road}, \text{background}\}$ in (\ref{focal_loss}) and accordingly $N$ in (\ref{focal_loss}) becomes 3.

\subsection{Objective Function}

Our objective function incorporates two losses as follows:
\begin{equation}
    L = \lambda_{1} \cdot L_{focal} + \lambda_{2} \cdot L_{dual-cycle},
\end{equation}
\noindent where $\lambda_{i} \in \{1, 2\}$ indicates the weight factor for each loss and we set $\lambda_{1}$ and $\lambda_{2}$ as 10 and $1 \cdot e^{-3}$, respectively. Note that the focal loss evaluates the discrepancy between the outputs of the DCT network (the outputs of the two decoders, i.e., the decoded $X'$) and the ground-truth road layout.


\section{Experiment}

To verify the performance of the proposed DCT network, we conduct a set of extensive experiments and compare the proposed model against multiple baselines in various scenarios.

\subsection{Settings}
\textbf{Datasets.} We employ two datasets for evaluation: the KITTI \cite{geiger2012we} and Argoverse \cite{chang2019argoverse} datasets. First, the KITTI dataset provides a number of splits. We evaluate the performance of each model on three KITTI splits: the KITTI 3D object detection split (KITTI 3D Object) \cite{chen2016monocular} with 3,712 training and 3,769 validation images for 3D vehicle detection, the KITTI Odometry split for road layout estimation with 15,806 training and 6,636 validation images, and the KITTI Raw \cite{schulter2018learning} with 10,156 training and 5,074 validation images for road layout estimation. Next, the Argoverse dataset consists of 6,723 training and 2,418 validation images for road layout estimation.

\textbf{Metrics.} For quantitative analysis, we adopt the standard object detection and segmentation metrics \cite{lin2014microsoft}: the mean of intersection-over-union (mIoU) and the mean average precision (mAP) metrics.

\textbf{Baselines.} We compare the performance of the proposed method against a set of state-of-the-art methods: MonoOccupancy \cite{lu2019monocular}, MonoLayout \cite{mani2020monolayout}, VPN \cite{pan2020cross}, Mono3D \cite{chen2016monocular}, OFTNet \cite{roddick2019orthographic}, and CVT \cite{yang2021projecting}. One thing to note is that we use Mono3D \cite{chen2016monocular} and  OFTNet \cite{roddick2019orthographic} for detecting vehicles in the top-view and adapted versions of VPN \cite{pan2020cross} and MonoOccupancy \cite{lu2019monocular} for fair comparison \cite{yang2021projecting}. Further, we utilize the latest performance of baselines reported online which is slightly distinctive than the reported results in their original papers. We exclude baselines that employ additional information such as multiple input images \cite{zhao2022jperceiver, saha2022translating} rather than a monocular image input.

\subsection{Implementation Details}
We implement the proposed model using the Pytorch framework and train the model using a single NVIDIA A30 GPU. We normalize the input images to 1024$\times$1024. The size of the ground truth label of KITTI Odometry, KITTI 3D Object, KITTI Raw and Argoverse are originally 128$\times$128, 256$\times$256, 256$\times$256, and 512$\times$512, respectively; they all get interpolated as 1024$\times$1024. Next, we render the ground truths of single-class instances as three-channel images by repeating the image content in the channel axis; we prioritize vehicles over roads to make each pixel contain one class label when ground truth pixels include multiple class labels. In addition, we render the ground truths of multi-class instances as one-hot encoded three-channel image. For feature extraction, we utilize the ResNet-18 architecture \cite{he2016deep} without bottleneck layers; for domain transfer between the front-view and the top-view, we employ MLPs with two fully-connected layers interspersed with the ReLU activation. For training, we employ the Adam optimizer \cite{kingma2015adam}. Moreover, the batch size, learning rate and the number of training epochs are 6, 1e-4 and 120, respectively. In the single class learning setting, we decay the learning rate once by 0.1 at the 50-th epoch; in the multi-class learning setting, we decay the learning rate once by 0.1 at the 100-th epoch. 
\subsection{Ablation Study}
To verify the effectiveness of each component of the proposed learning scheme, we conduct ablation studies in two learning settings: single-class learning and multi-class learning.

\textbf{Single-Class Learning.} Table \ref{tb:result_ablation_singleclass} describes the ablation study result in the single-class learning scenario. The baseline model learned with the naive cross-entropy loss. As we applied the focal loss to the baseline model, the performance enhanced since the proposed learning scheme with the focal loss effectively dealt with the class imbalance. Moreover, the dual cycle loss further improved the performance as the loss incorporated the complete dual cycle consistency. In short, sequential application of the focal loss and the proposed dual cycle loss to the baseline model boosts the performance sequentially---which demonstrates the effectiveness of both the proposed application of the focal loss and the designed dual cycle loss.

\textbf{Multi-Class Learning.}
Table \ref{tb:result_ablation_multiclass} describes the ablation study result in the multi-class learning scenario. Similar to the single-class learning scenario, the application of the focal loss and the dual-cycle loss in progression to the baseline model enhanced the performance with substantial performance margins consecutively---corroborating the validity of each component of the proposed learning scheme in the multi-class learning setting. However, note that the multi-class learning deteriorates performance and the performance drop is more severe for vehicles. We surmise that vehicles occupying less areas than roads in a road layout require more deliberate learning scheme design.

\begin{table}[!t]
\caption{Ablation study results using different losses in the single-class learning scenario on the Argoverse Dataset.}
\label{tb:result_ablation_singleclass}
\begin{center}
\begin{footnotesize}
\begin{tabular}{ccccc}
\toprule
\multicolumn{1}{c}{\multirow{2}[2]{*}{\textbf{Loss items}}} & \multicolumn{2}{c}{\textbf{Argoverse Road}} & \multicolumn{2}{c}{\textbf{Argoverse Vehicle}} \\ 
\cmidrule(lr){2-5}
 & mIoU (\%) & mAP (\%) & mIoU (\%) & mAP (\%)  \\

\midrule
Baseline & 76.56& 87.30 & 47.86 & 62.69 \\ 
\midrule
Focal &\textbf{76.74} &88.47  & 47.41 & 67.48\\ 
Focal +Dual Cycle &76.71&\textbf{88.86}&\textbf{47.94}&\textbf{68.95}\\
\bottomrule
\end{tabular}
\end{footnotesize}
\end{center}
\end{table}

\begin{table}[!t]
\caption{Ablation study results using different losses in the multi-class learning sceneario on the Argoverse Dataset.}
\label{tb:result_ablation_multiclass}
\begin{center}
\begin{footnotesize}
\begin{tabular}{ccccc}
\toprule
\multicolumn{1}{c}{\multirow{2}[2]{*}{\textbf{Loss items}}} & \multicolumn{2}{c}{\textbf{Argoverse Road}} & \multicolumn{2}{c}{\textbf{Argoverse Vehicle}} \\ 
\cmidrule(lr){2-5}
 & mIoU (\%) & mAP (\%) & mIoU (\%) & mAP (\%)  \\

\midrule
Baseline& 74.45 & 86.51 & 29.53 & 43.84 \\  
\midrule
Focal &73.93 & 86.00 & \textbf{31.89} & 46.19\\
Focal +Dual Cycle  &\textbf{74.73}& \textbf{86.76} &31.75&\textbf{46.20}\\
\bottomrule
\end{tabular}
\end{footnotesize}
\end{center}
\end{table}

\subsection{Comparative Study}
\begin{figure*}[t]
\begin{center}
   \includegraphics[width=1\linewidth]{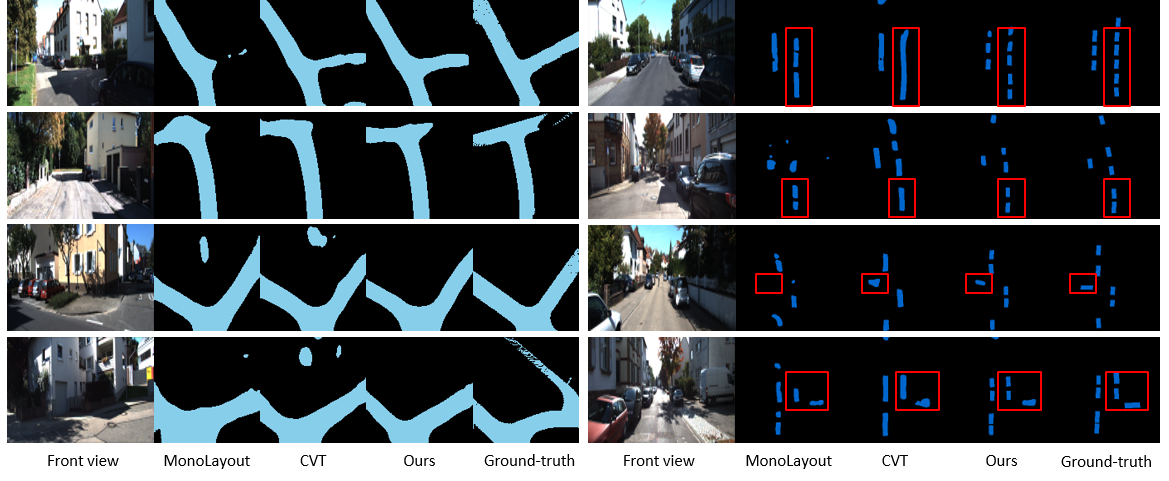}
\end{center}
   \caption{The qualitative comparison results of road layout estimation and vehicle occupancy segmentation in BEV on the KITTI Odometry dataset and the KITTI 3D Object dataset, respectively. In short, the proposed DCT network displays superior performance compared to the latest baseline models in both tasks; the DCT network can transform the road layout from the front-view to the top-view more accurately and even distinguish adjacent vehicles in a row.}
\label{fig:qual}
\end{figure*}
\begin{figure}[t]
\begin{center}
   \includegraphics[width=0.95\linewidth]{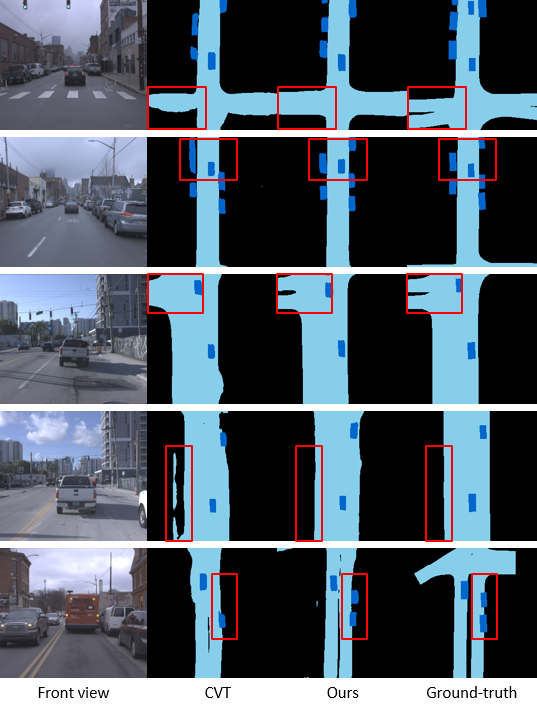}
\end{center}
   \caption{The qualitative comparison results of simultaneous road layout estimation and vehicle occupancy segmentation in BEV on the Argoverse dataset. The proposed DCT network displays less noise and produces more accurate segmentation results compared to the baseline method.}
\label{fig:multiclass_qual}
\end{figure}

We perform comparative studies to establish the performance contribution against the baseline models. Note that, in Tables \ref{tb:result_comparative} and \ref{tb:result_comparative_3d}, * means the latest performance published online, which are slightly different from the reported results in the original papers.

\textbf{Single-Class Learning.}
Tables \ref{tb:result_comparative} and \ref{tb:result_comparative_3d} exhibit the comparative study result in the single-class learning scenario. The proposed DCT architecture shows state-of-the-art performance on both the KITTI and Argoverse datasets with substantial performance gains. The performance gain is more prominent in mAP than mIoU (7.85\% and 6.26\% mAP enhancement in the KITTI 3D and Argoverse vehicle datasets, respectively) since we proposed to handle the class imbalance utilizing the focal loss.

Moreover, Fig. \ref{fig:qual} presents the qualitative comparison of our approach against the baseline methods. The comparison result demonstrates that our method overcomes existing class imbalance problem effectively. The predicted road layout from the DCT network contains less noise than the baselines (in fact, almost zero noise). Plus, the DCT network generates road layouts with more accurate and clearer boundaries. Next, our DCT method shows superior performance in the 3D object detection in BEV task. The DCT network successfully distinguishes densely arranged cars in a row (marked as red boxes in the figure) whereas the baseline models likely failed to differentiate them.

\textbf{Multi-Class Learning.}
Table \ref{tb:result_comparative_multiclass} quantifies the comparative study result in the multi-class learning scenario; the proposed DCT network accomplished state-of-the-art performance with considerable margins. The performance margin is more conspicuous for predicting vehicles where the class imbalance problem is more severe. In addition, all methods considered in the experiment displayed degraded performance in the multi-class learning scenario compared to the single-class learning scenario. We assume that vehicles taking up much fewer pixels than roads in a road layout cause the performance degradation---demanding an advanced learning scheme for multi-class learning.

Besides, Fig. \ref{fig:multiclass_qual} illustrates the qualitative comparison result. The qualitative comparison result establishes the superiority of the proposed DCT architecture likewise. Our DCT architecture transforms the road layout from the front-view to the top-view with more accurate and clearer boundaries compared to the baseline model; furthermore, the road layouts from the DCT network contains less noise. In addition, the DCT network differentiates adjacent vehicles, which the baseline model can hardly distinguish.

\begin{table*}[!t]
\caption{Comparative study results of top-view semantic segmentation on the KITTI and Argoverse datasets.}
\label{tb:result_comparative}
\begin{center}
\begin{small}
\begin{tabular}{ccccccccc}
\toprule
\multicolumn{1}{c}{\multirow{2}[2]{*}{\textbf{Method}}} & \multicolumn{2}{c}{\textbf{KITTI Raw}} & \multicolumn{2}{c}{\textbf{KITTI Odometry}} & \multicolumn{2}{c}{\textbf{Argoverse Road}} & \multicolumn{2}{c}{\textbf{Argoverse Vehicle}} \\ 
\cmidrule(lr){2-5}
\cmidrule(lr){6-9}
 & mIoU (\%) & mAP (\%) & mIoU (\%) & mAP (\%) & mIoU (\%) & mAP (\%) & mIoU (\%) & mAP (\%) \\

\midrule

MonoOccupancy \cite{lu2019monocular} & 58.41 & 66.01 & 65.74 & 67.84 & 72.84 & 78.11 & 24.16 & 36.83\\
VPN \cite{pan2020cross} & 59.58 & 79.07 & 66.81 & 81.79 & 71.07 & 86.83 & 16.58 & 39.73 \\
MonoLayout \cite{mani2020monolayout} & \textbf{66.02} & 75.73 & 76.15 & 85.25 & 73.25 & 84.56 & 32.58 & 51.06\\
CVT$^{*}$ \cite{yang2021projecting} & 64.13 & 83.37 & \textbf{77.47} & 86.39 & 76.56 & 87.30 & 47.86 & 62.69\\
\midrule

\textbf{Ours}& 65.86&  \textbf{86.56}&  77.15& \textbf{88.28} & \textbf{76.71}&  \textbf{88.86}& \textbf{47.94} & \textbf{68.95}\\

\bottomrule
\end{tabular}
\end{small}
\end{center}
\end{table*}

\begin{table}[!t]
\caption{Comparative study results on the KITTI 3D Object dataset.}
\label{tb:result_comparative_3d}
\begin{center}
\begin{small}
\begin{tabular}{ccc}
\toprule
\multicolumn{1}{c}{\multirow{2}[2]{*}{\textbf{Method}}} & \multicolumn{2}{c}{\textbf{KITTI 3D Object}}\\ 
\cmidrule(lr){2-3}
 & mIoU (\%) & mAP (\%) \\

\midrule

MonoOccupancy \cite{lu2019monocular} & 20.45 & 22.59\\
Mono3D \cite{chen2016monocular}& 17.11 & 26.62\\
OFTNet \cite{roddick2019orthographic} & 25.24 & 34.69\\
VPN \cite{pan2020cross}& 16.80 & 35.54\\
MonoLayout \cite{mani2020monolayout}& 30.18 & 45.91\\
CVT$^{*}$ \cite{yang2021projecting} & 38.85 & 51.04\\ 
\midrule

\textbf{Ours}& \textbf{39.44} & \textbf{58.89} \\

\bottomrule
\end{tabular}
\end{small}
\end{center}
\end{table}

\begin{table}[!t]
\caption{Comparative study results of Multi-Class learning on the Argoverse dataset.}
\label{tb:result_comparative_multiclass}
\begin{center}
\begin{footnotesize}
\begin{tabular}{ccccc}
\toprule
\multicolumn{1}{c}{\multirow{2}[2]{*}{\textbf{Method}}} & \multicolumn{2}{c}{\textbf{Argoverse Road}} & \multicolumn{2}{c}{\textbf{Argoverse Vehicle}} \\ 
\cmidrule(lr){2-5}
 & mIoU (\%) & mAP (\%) & mIoU (\%) & mAP (\%)  \\

\midrule
MonoLayout \cite{mani2020monolayout} & 67.29& 79.57 & 16.39 & 26.42 \\
CVT \cite{yang2021projecting} & 74.40 & \textbf{87.07} & 30.02& 44.07 \\
\midrule
\textbf{Ours} &\textbf{74.73}&86.76&\textbf{31.75}&\textbf{46.20}\\

\bottomrule
\end{tabular}
\end{footnotesize}
\end{center}
\end{table}

\section{Conclusion}
In this work, we proposed the dual-cycled cross-view transformer (DCT) network for unified road layout estimation and 3D object detection. The DCT network concurrently performs both the road layout estimation and 3D object detection tasks in BEV, which are essential for autonomous driving. We proposed to manage the class imbalance of the training data through the focal loss so that excessive negative samples would not overwhelm the learning process. Besides, we designed dual cycle consistency loss to further offset the performance loss caused by the class imbalance. Next, we investigated the effect of multi-class learning on road layout estimation; otherwise we need one neural network per each class. We revealed that multi-class learning requires an enhanced model design since the current model's learning capacity does not satisfy the complexity of the task. To establish the effectiveness of the DCT architecture, we conducted a set of comprehensive experiments. The experiment results corroborate the effectiveness of the proposed model and we obtained state-of-the-art performance on both semantic raod layout estimation and 3D object detection benchmark datasets. The source code and the pretrained mode weights are open-source.


\bibliographystyle{IEEEtran}
\bibliography{main}

\end{document}